\documentclass[10pt, conference, compsocconf]{IEEEtran}

\usepackage[cmex10]{amsmath}
\usepackage{amssymb}
\usepackage[dvips]{graphicx}
\usepackage{qtree}
\usepackage[turnoff]{notes}
\usepackage{multirow}

\DeclareMathOperator{\mysim}{\omega}
\DeclareMathOperator{\acc}{\text{t5p}}
\DeclareMathOperator{\diff}{\delta}
\DeclareMathOperator{\mae}{\text{mae}}

\newcommand{\vect}[1]{\boldsymbol{#1}}

\makeatletter
\newcommand{\subsumeLines}[1]{
\begin{picture}(0,1)
\put(-0.1,0){\line(0,1){1}}
\put(0.1,0){\line(0,1){1}}
\end{picture}
}
\newcommand{\subsumeLinesLeft}[1]{
\begin{picture}(2,0.5)
\put(-0.05,0){\line(2,1){1}}
\put(0.05,0){\line(2,1){1}}
\put(2,0){\line(-2,1){1}}
\end{picture}
}
\newcommand{\instanceLines}[1]{
\begin{picture}(0,1)
\linethickness{0.4mm}
\put(0,0){\line(0,1){1}}
\end{picture}
}
\let\qdrawReal=\qdraw@branches
\newcommand\brSubsume{\let\qdraw@branches=\subsumeLines}
\newcommand\brSubsumeLeft{\let\qdraw@branches=\subsumeLinesLeft}
\newcommand\brInstance{\let\qdraw@branches=\instanceLines}
\newcommand\brRestore{\let\qdraw@branches=\qdrawReal}
\makeatother

\begin{document}
\title{Learning Heterogeneous Similarity Measures for Hybrid-Recommendations in Meta-Mining}
\author{\IEEEauthorblockA{Phong Nguyen, Jun Wang and Melanie Hilario} 
\IEEEauthorblockA{AI group, Department of Computer Science\\
CUI, University of Geneva\\
Geneva, Switzerland\\
Phong.Nguyen@unige.ch}
\and \IEEEauthorblockN{Alexandros Kalousis}
\IEEEauthorblockN{ Department of Business Informatics\\
University of Applied Sciences\\
Western Switzerland\\
Alexandros.Kalousis@unige.ch}
}
\maketitle

\begin{abstract}

The notion of meta-mining has appeared recently and extends the traditional meta-learning in two ways. First it does
not learn meta-models that provide support only for the learning algorithm selection task but ones that support
the whole data-mining process. In addition it abandons the so called black-box approach to algorithm description followed 
in meta-learning.  Now in addition to the datasets, algorithms also have descriptors, workflows as well. For the latter two 
these descriptions are semantic, describing properties of the algorithms, such as cost functions, learning biases, etc. With
the availability of descriptors both for the datasets and the data-mining workflows the traditional modelling techniques 
followed in meta-learning, typically based on classification and regression algorithms, are
no longer appropriate. Instead we are faced with a problem the nature of which is much more similar to the problems that appear
in recommendation systems. However on the same time the requirements of the meta-mining tasks make the direct use of tools from
recommender systems rather inappropriate. The most important meta-mining requirements are that suggestions should use only 
the datasets and workflows descriptors and the cold-start problem, e.g. providing workflow suggestions for new datasets. 

In this paper we take a different view on the meta-mining modelling problem and treat it as a recommender problem. In order 
to account for the meta-mining specificities we derive a novel metric-based-learning recommender approach. Our method learns 
two homogeneous metrics, one in the dataset and one in the workflow space, and a heterogeneous one in the dataset-workflow space.  
All learned metrics reflect similarities established from the dataset-workflow preference matrix. The latter is constructed 
from the performance results obtained by the application of workflows to datasets. We demonstrate our method on meta-mining 
over biological (microarray datasets) problems. The application of our method is not limited to the meta-mining problem, 
its formulations is general enough so that it can be applied on problems with similar requirements.

\note[Removed]{Similarity/distances measures play a central role in meta-learning.  Being able to situate appropriately a new dataset in the algorithm's 
performance space in order to retrieve the best algorithms to apply has been the main concern in the meta-learning community during these 
last decades.  However, previous similarity/distances measures proposed for meta-learning have been mainly defined with the standard Euclidean 
distance which is limited by assuming as equally informative the data features used to build these measures.  Moreover, meta-learning as originally 
defined considers algorithms as a black-box and focus exclusively on the learning algorithm without considering others steps in the knowledge discovery process, 
such as data cleaning, data selection, feature extraction and selection, etc.  In this work, we propose a metric-learning approach for meta-mining. 
Meta-mining is a subclass of meta-learning which tears open the algorithm's black-box and defines algorithm/workflow features according to a declarative 
representation of the data mining process, the Data Mining OPtimization ontology. Our goal is to learn robust similarity measures for datasets and 
algorithms with respect to their features and the algorithm's performance space.  Additionally, we propose to learn in a single approach these 
similarity measures in order to make them heterogenous, i.e. similarity between datasets is related to similarity between algorithms and vice-versa. 
We show in this work that first using metric-learning is an obvious way to learn appropriate similarity measures for meta-mining, 
and second correlating our similarity measures may also improve meta-mining and meta-learning in general.}

\end{abstract}

\begin{IEEEkeywords}
Meta-Mining; Meta-Learning; Hybrid Recommendation; Metric-Learning; 
\end{IEEEkeywords}

\section{Introduction}

Meta-learning is learning to learn: in computer science, it is the application of machine learning techniques to meta-data describing past learning experience, 
typically applications of learning algorithms to specific datasets, in order to derive meta-learning models that can support the selection of an appropriate
algorithm for a new dataset, ~\cite{Giraud-Carrier2004,Anderson2007,Brazdil2008,Vilalta_Giraud-Carrier_Brazdil_Soares_2004}. The meta-learning models are usually
classification or regression models learned by standard classification and regression algorithms. Until very recently meta-learning
was focusing only on the learning part of the data mining process, by trying to model the behavior of different learning algorithms, and was treating the learning
algorithms as black-boxes making no effort to describe the concepts that underline them and their properties. 

The authors of \cite{Hilario2010} made an effort to address these
limitations by extending the meta-learning process to the whole data mining process  resulting in a more comprehensive task which they called {\em meta-mining}.
In addition they made use of a data mining ontology in order to provide detailed descriptions of data mining algorithms in terms of their core components, 
underlying assumptions, cost functions, optimization strategies, etc, as well as detailed descriptions of data mining workflows, the latter
composed of operators implementing data mining algorithms. Even though the introduction of data mining algorithm and workflow descriptors was
an important step the authors made rather poor use of them by modelling the meta-mining problem as a classification problem, following thus
the traditional meta-learning modelling approach. In this classification
problem the meta-mining instances corresponded to data mining experiments, applications of workflows or algorithms on datasets, and they consisted 
of two types of features, features that described the dataset and features that describe the data mining workflow. The class label was determined on the basis of the
performance result estimated by the application of the workflow on the dataset and was indicating the appropriateness or not of the workflow for the
dataset.

In this paper we take a different approach on the modelling of the meta-learning and meta-mining tasks. We view them as a matching problem 
between datasets on the one hand and data mining algorithms or workflows on the other, in which the matching criterion is the performance 
of the latter when applied on the former. We will address three different meta-mining tasks.
Given a new dataset we want to recommend or rank available algorithms or data mining workflows in terms
of their expected performance on the specific dataset; we will call this task {\em learning workflow preferences}. Symmetrically to 
this we want, given a new data mining workflow or algorithm, to know for which datasets they are most appropriate; we will call 
this task {\em learning dataset preferences}. Finally, given a new dataset and a new workflow or algorithm we want to be able to determine
the goodness of their match, i.e. the degree to which the latter will have a good performance when applied to the former; we will call this 
{\em learning dataset-workflow preferences}. It is obvious that all these should be determined without any actual application of the new algorithms 
on the new datasets but on the basis of some meta-mining model that will be learned from the past mining experiences.

These type of problems are similar in nature to problems that appear in recommender systems, where we have users and items and we want 
to suggest additional items for a given user based on the preferences of users with similar preferences. In the meta-mining and meta-learning 
case the matrix containing the preferences of users for items is replaced 
by a performance based matrix of datasets and workflows or algorithms that indicates the performance of the latter applied to the former. 
This performance-based preference matrix will be one component of our meta-mining data; in addition we will use dataset and workflow 
or algorithm descriptors. The final meta-mining models will only use the dataset and workflow descriptors to return the preferences.
In recommender systems there is a relevant stream of work that makes use of descriptors of users and items, similar to the 
descriptors of datasets and workflows, that is called hybrid recommendation systems~\cite{Agarwal2010,Agarwal2009,Stern2009,Burke2002}. 
\note[Alexandros]{Is there some more specific reference and/or more recent reference for hybrid recommendation systems?}
However there are also a number of differences between the nature of the recommendation problem that we have in meta-mining
and the typical recommendation problems. In the latter the preferences matrix is often very sparse, of 
high dimensionality, and can have hundreds of thousands of rows/users. In contrast, the preferences
matrix in meta-mining is rather dense and involves few hundreds of datasets and workflows.
The features of datasets and workflows that we use in the meta mining problems are quite informative
in contrast to the typical recommendation problems where it is rather hard to get informative features 
especially in what concerns user descriptions. Finally, in the meta mining setting, 
the cold-start problem is central, with the most typical example being predicting the workflow preferences 
for a new dataset. However, in recommendation problem, partly due to the low information content of the features
describing users, but also due to the nature of the problem itself, the main focus is in the completion of missing 
values in the preferences matrix based on historical ratings of items by similar users.

In this paper we present a new metric-learning-based approach to hybrid recommendation for meta-mining, 
which learns to match dataset descriptors to workflow descriptors. More specifically we will learn three 
different metrics.  One on the dataset descriptor space which will reflect the 
fact that similar datasets will have similar workflow preferences, as these are given by the performance-based preference 
matrix. One on the workflow descriptor space which will reflect the fact that similar workflows will have similar dataset preferences again as these are given
by the performance-based preference matrix. And a last heterogeneous metric over the two spaces of dataset 
and workflow descriptors, which will directly give the similarity/appropriateness of a given dataset for 
a given workflow. We will use these learned metrics, alone or in combination, to address the three meta-mining 
tasks that we described in the previous paragraphs. 

To the best of our knowledge the metric learning approach that we present is the first of its kind, 
not only for meta-mining, but also for the general context of hybrid recommendation problems.
Even though it was developed to address the specific requirements of the meta-mining setting
it is not specific to it, and it can be used in any kind of recommendation system that has 
similar requirements, i.e. preference based matchings of users to items based on descriptions of them and cold-start problem. 

\note[Removed]{
As originally defined applies specifically to learning, 
which is only one \textemdash{} albeit the central \textemdash{} step in the data mining (DM) process. We still lack an understanding of how the different 
components of the DM process (e.g. data cleaning, data selection, feature extraction and selection, model pruning, and model aggregation) interact; there 
are no clear guidelines except for high-level process models such as \textsc{crisp-dm}~\cite{crisp}.

In order to fill these gaps,~\cite{Hilario2010} proposed a new meta-learning framework which: 1) extends meta-learning to the full knowledge discovery 
process, thus subsuming classical meta-learning under a more comprehensive approach called {\em meta-mining}; 2) opens the black box of algorithms and 
analyses them in terms of their core components such as their underlying assumptions, cost functions, and optimization strategies; 3) grounds meta-mining 
on a declarative representation of the DM process, such as that used in the Data Mining OPtimization ontology (DMOP) %
\footnote{The DMOP is available at http://www.dmo-foundry.org%
}. Whereas previous meta-learning was based on the Rice model~{[}Rice1976{]} which relied exclusively on problem/data features to select algorithms, 
the proposed framework takes algorithm features into account and defines an extended Rice model as follows.

{\em Let $\mathcal{X}$ be the space of learning problems (or datasets) and $\mathcal{A}$ the space of learning algorithms (or workflows), both characterized by the feature spaces $\mathcal{F}$ and $\mathcal{G}$ respectively. Further, let $\mathcal{P}$ be the performance space that we want to observe. For a given dataset $x\in\mathcal{X}$ characterized by $f(x)\in\mathcal{F}$ and workflows $a\in\mathcal{A}$ characterized by $g(a)\in\mathcal{G}$, find one (or more) workflows $\{a\}$ via the selection mapping $S(f(x),g(a))$ such that the performance mapping $p(a(x))\in\mathcal{P}$ is maximized.}


Meta-mining thus addresses the {\em algorithm selection task}~\cite{Smith-Miles_2008} in a novel manner: first, it considers the full DM workflow 
process rather than the learning phase alone; second, it redefines the selection mapping $S$ by \emph{combining} commonly used problem/dataset features 
with semantic features of algorithms/workflows as defined in the DMOP ontology.  Additionally, since $S$ is symmetric, meta-mining can also address the 
task of selecting the right learning datasets $\{x\}\in\mathcal{X}$ for a given workflow $a\in\mathcal{A}$. 
Generally speaking, both the algorithm/workflow and problem/dataset selection tasks can be formalized as the inference of the {\em related} 
performance $p(a(x))$ for a given dataset $x\in\mathcal{X}$ and a given workflow $a\in\mathcal{A}$ in the performance space $\mathcal{P}$ with respect 
to their meta-features $f(x)\in\mathcal{F}$ and $g(a)\in\mathcal{G}$.
}

\note[Removed]{
In this paper, we propose to tackle meta-mining as a hybrid recommandation problem~\cite{Burke2002}; 
whereas in~\cite{Hilario2010}, the authors considered meta-mining as a classification problem, we consider 
the inference of the performance space $\mathcal{P}$ as a learning to rank problem in the meta-feature spaces $\mathcal{F}$ and $\mathcal{G}$ 
such that we will be able to solve the well-known {\em cold-start} problem for new datasets and new workflows in the same time.
We will address our meta-mining tasks with a metric-learning approach~\cite{xing2003dml} 
from which in a first step distinct (homogenous) similarity measures are learned for each of the respective meta-spaces in order to address our selection tasks separately.
Then, more importantly, we will project the $\mathcal{F}$ and $\mathcal{G}$ spaces into a common space associated with the $\mathcal{P}$ space, such that (heterogenous) similarity measures defined for datasets and workflows cohere in a single approach to our selection tasks. In short, the goal is to find meaningful similarity measures that will allow us to compute an optimal selection mapping $S(f(x),g(a))$ following memory-based meta-learning approaches {\cite{Soares2000,Hilario2001}.}
}

The rest of the paper is organized as follows. In section \ref{sec:task}, we define the meta-mining tasks.
In section \ref{sec:metric}, we describe our metric-learning based approach to the problem of learning hybrid recommendations for meta-mining. 
In section \ref{sec:metadata}, we present briefly the characteristics---features--- that we use to describe the datasets and the workflows. 
In section \ref{sec:exp}, we give the experiments and the evaluation of our approach. In section \ref{sec:rworks}, we discuss the related work and
finally we conclude in section \ref{sec:conc}.

\section{Meta-Mining Tasks}
\label{sec:task}
Before proceeding to the definition of the different meta-mining tasks that will address let us give some 
necessary notations. Let $\mathbf x=(x_1,\dots x_d)^T \in \mathbb R^d$ be the description of some dataset, and 
$\mathbf X$ an $n \times d$ dataset matrix the $i$th row of which is given by the $\mathbf x_i^T$ dataset. Thus 
the $\mathbf X$ matrix is the set of datasets over which the meta-mining will take place. In addition let
$\mathbf a=(a_1, \dots, a_l)^T \in \mathbb R^l$  be the description of some data mining workflow, and $\mathbf A$ an
$m \times l$ workflow matrix the $j$th row of which is the $\mathbf a_j^T$ workflow, i.e. $\mathbf A$ will be the 
data mining workflow matrix over which the meta-mining will take place. Finally let $\mathbf R$ be an $n \times m$ matrix
the $(i,j)$ entry of which depends on some performance result obtained by the application of the $\mathbf a_j$ data mining
workflow on the $\mathbf x_i$ dataset. We will use the notation $\mathbf r_{\mathbf x_i}$ to denote the vector given by the 
row of $\mathbf R$ which corresponds to the $\mathbf x_i$ dataset and which contains the performance measures obtained by the 
application of the $m$ data mining workflows on $\mathbf x_i$, and the notation $\mathbf r_{\mathbf a_j}$ to denote the vector given by the 
$j$th column of $\mathbf R$ which contains the performance results of the application of the $a_j$ data-mining workflow
on the $n$ datasets. Thus the $\mathbf R$ matrix relates, based on performance, datasets with workflows and can be seen 
as giving the appropriateness of workflows for datasets and vice versa.

Since here we will focus only on meta-mining for classification problems the performance measure that we will be 
using to fill up $\mathbf R$ will be based on classification accuracy which we will estimate by ten-fold cross-validation. 
The accuracies achieved by different workflows are not comparable over different datasets, what is 
much more important in meta-learning and meta-mining is the relative performance order of a set of 
data mining workflows or algorithms on a given dataset; this relative order can be compared in a 
meaningful manner over different datasets. We devise such a relative order in the following way. Given 
a pair of classification data mining workflows $\mathbf a_k$ and $\mathbf a_l$ applied on dataset $\mathbf x_i$
we compute the statistical significance of their accuracies differences using a McNemar's test, with a p-value of 0.05.
If one workflow is statistically significant better than the other it is assigned a score of one and the other a score 
of zero, in case of no significant difference both are assigned a score of 0.5. For a given dataset $\mathbf x_i$ 
the score of a workflow $\mathbf a_k$ will be the sum of the points it gets in all its pairwise comparisons with the
other $m-1$ workflows. It is this score that we will use to populate
the $\mathbf R$ matrix, i.e. its $(i,j)$ entry will be the score obtained by workflow $\mathbf a_j$ on dataset $\mathbf x_i$, 
we will also use the notation $r_{\mathbf x_i, \mathbf a_j}$ to denote the $(i,j)$ entry of $\mathbf R$.

Given the above we will now define three different meta-mining tasks. In the first one given a new unseen dataset $\mathbf x$, 
i.e. a dataset with which we have not experimented with, we want to estimate the relative performance order of the $m$ 
data mining workflows. In other words we want to estimate the relative workflow performance, or workflow preference,  
vector $\mathbf r_{\mathbf x}$ for the $\mathbf x$ dataset. We will call this task {\em learning workflow preferences}.
The second meta-mining task is the symmetric of the first; here we want to estimate appropriateness
of a new unseen workflow $\mathbf a$ for the $n$ datasets, i.e. we want to estimate the dataset preference 
vector $\mathbf r_{\mathbf a}$ for the $\mathbf a$ workflow. We will call this task {\em learning dataset preferences},
Finally in the third, last and most difficult, meta-mining task we want to estimate the appropriatness of an unseen workflow 
$\mathbf a$ on an unseen dataset $\mathbf x$, i.e. estimate the $r_{\mathbf x,\mathbf a}$ value. We will call this
meta-mining task {\em learning dataset-workflow preferences}.

To address all three tasks we will rely on the use of appropriate similarity measures. To learn workflow preferences 
we will need a dataset similarity measure that given a new dataset $\mathbf x$ will establish its most similar 
datasets in the training set $\mathbf X$. From the workflow preference vectors of these datasets we will then 
estimate the workflow preference vector $\mathbf r_{\mathbf x}$ of $\mathbf x$. In the same manner to learn dataset 
preferences we need a workflow similarity measure that given a new workflow $\mathbf a$ will establish its most similar workflows
in the training set $\mathbf A$. From the dataset preference vectors of these workflows we will then estimate the dataset
preference vector $\mathbf r_{\mathbf a}$ of $\mathbf a$. For the last task we will rely on the use of an {\em heterogeneous}
similarity measure that computes directly similarities between workflows and datasets, which thus given an unseen
dataset $\mathbf x$ and an unseen workflow $\mathbf a$ will produce the $r_{\mathbf x, \mathbf a}$ corresponding
to the appropriateness of $\mathbf a$ for $\mathbf x$.

In the following section we will show how to learn appropriate metric matrices that we will use to compute the similarity
measures that we briefly described in the previous paragraph.

\section{Learning Similarities for Hybrid-Recommendations in Meta-Mining}
\label{sec:metric}
Before starting to describe in detail how we will address the three meta-mining tasks
let us take a step back and give a more abstract picture of the
type of learning setting that we want to address. We have two types of learning instances, $\mathbf x \in \mathcal X$, and $\vect a \in \mathcal A$,
and two training matrices $\mathbf X: n \times d$ and $\mathbf A: m \times l$ respectively. Additionally we also have an instance alignment or 
preference matrix $\mathbf R: n \times m$, the $R_{ij}$ entry of which gives some measure of appropriateness, preference, or match of the 
$\mathbf x_i$ and $\vect a_j$ instances. 

We can construct a similarity matrix for the instances of the $\mathbf X$ by exploiting the idea that similar instances of the $\mathbf X$
should have similar preferences with respect to the instances of the $\mathbf A$ matrix. Here we do not rely anymore in the original representation
of the $\mathbf x$ instances in order to define their similarities but on their preferences with respect to the $\mathbf a$ instances\footnote{This 
reflects one of the basic assumptions
in metalearning, the fact that what we are trying to reflect is a similarity of datasets in terms of the relative performance/appropriateness
of different learning paradigms/algorithms for them}. So the $\mathbf x$ instances similarity matrix will be the $\mathbf R \mathbf R^T$ 
matrix, the $[\mathbf{R} \mathbf{R}^\text{T}]_{ij}$ entry of which will give the similarity of the $\mathbf x_i$ and $\mathbf x_j$ instances.
In exactly the same manner we can construct the similarity matrix for the $\mathbf a$ instances as $\mathbf R^T \mathbf R$. 

We now want to learn two Mahalanobis metrics one in the  $\mathcal X$ and one in the  $\mathcal A$ space which 
will reflect the instance similarities as these are given by the $\mathbf R \mathbf R^T$ and $\mathbf R^T \mathbf R$ 
similarity matrices respectively. In addition we want to learn a third metric over the two heterogeneous spaces  
$\mathcal X$ and $\mathcal A$ which will reflect the similarity/preference of an $\mathbf x_i \in \mathbf X$ 
instance to an $\mathbf a_j \in A$ instance as this is given by the $R_{ij}$  preference value. Since learning
a Mahalanobis metric is equivalent to learning a linear transformation we will see in the following paragraphs 
that what we actually need to learn is eventually two such linear transformations, one for the $\mathcal X$ and 
one for the  $\mathcal A$ space, which will optimize the three objective functions that we just sketched.

We should note here that the setting that we just described is not specific to the meta-mining context but is
also relevant for any recommendation problem with similar requirements. To the best of our knowledge the metric-based 
solution which we will present right away is the first of its kind for such settings.

 
\subsection{Learning a dataset metric}
\label{sec:dsim}

We will now describe how to learn a Mahalonobis metric matrix $\mathbf W_{\mathcal X}$ in the $\mathcal X$ dataset space in a manner
that will reflect datasets similarity in terms of the similarity of their workflow preference vectors.
Instead of using the $\mathbf{R} \mathbf{R}^\text{T}$ matrix to establish the similarity of two datasets in terms of their 
preference vectors, under which the dataset similarity is simply the inner product of the workflow preference vectors,
we will rely on the Pearson rank correlation coefficient of these preference vectors.  The latter is a more 
appropriate measure of dataset similarity since it focuses on the relative workflow performance which is more relevant
when one wants to measure dataset similarity. Nevertheless to simplify notation we will continue using the 
 $\mathbf{R} \mathbf{R}^\text{T}$ notation. 

We define the following metric learning optimization problem: 
\begin{eqnarray*}
\label{eq:metric-dataset.1}
\min_{{\mathbf W}_{\mathcal X}} F_1(\mathbf W_{\mathcal X})
     & = & ||\mathbf{R} \mathbf{R}^\text{T} - \mathbf{X} \mathbf{W}_{\mathcal X} \mathbf{X}^\text{T}||_F^2 + \mu_1 tr(\mathbf{W}_{\mathcal X}) \\ \nonumber
s.t. &   & \mathbf W_{\mathcal X} \succeq 0 \nonumber
\end{eqnarray*} 
where $||.||_F$ is the Frobenius matrix norm, $tr(.)$ the matrix trace, and $\mu_1 \geq 0$ is a parameter controlling the trade-off 
between empirical error and the metric complexity used to control overfitting, which is a convex optimization problem. As already mentioned 
learning a Mahalanobis metric matrix is equivalent to learning a linear transformation of the original feature space. Thus we can now rewrite 
our metric learning problem with the help of a linear transformation as:
\begin{eqnarray}
\label{eq:metric-dataset.2}
\min_{\mathbf U} F_1(\mathbf U)
     & = & ||\mathbf{R} \mathbf{R}^\text{T} - \mathbf{X} \mathbf{U} \mathbf U^T \mathbf{X}^\text{T}||_F^2 + \mu_1 ||\mathbf{U}||_F^2 
\end{eqnarray} 
where $\mathbf W_{\mathcal X}= \mathbf U \mathbf U^T$ is the $d \times d$ 
metric matrix, and $\mathbf U$ an associated linear transformation with 
dimensionality $d \times t$ which projects the dataset description to 
a new space of dimensionality $t$.  Unlike the previous optimization problem 
this is no longer convex. We will work with optimization problem 
(\ref{eq:metric-dataset.2}) because it will make easier the variable sharing 
between the different optimization problems that we will define. We solve 
it using gradient descent.

Using the learned metric the similarity of two datasets $\mathbf x_i$ and $\mathbf x_j$ is  
$\mysim(\mathbf x_i, \mathbf x_j) = \mathbf x_i \mathbf U \mathbf U^T \mathbf x_j$. Given some
new dataset $\mathbf x$ we will use this similarity to establish the set $N_{\mathbf x}$ consisting of 
the $N$ datasets that are most similar to $\mathbf x$ with respect to the similarity of their 
relative workflow preferences as this is computed in the original feature space $\mathcal X$. 
With the help of $N_{\mathbf x}$ we can now compute the workflow preference vector of $\mathbf x$
as the weighted average of the workflow preference vectors of its nearest neighbors by: 
\begin{eqnarray}
\mathbf r_{\mathbf x} = \zeta_{\mathbf x}^{-1} \sum_{x_i \in N_{x}} \mathbf r_{\mathbf x_i} \mysim_{\mathcal X} (\mathbf x,\mathbf x_i)
\end{eqnarray}
where $\zeta_{\mathbf x}$ is a normalization factor given by $\zeta_{\mathbf x} = \sum_{\mathbf x_i \in X} \mysim_{\mathcal X} (\mathbf x, \mathbf x_i)$.
Thus using the learned metric we can compute the workflow preference vector $\mathbf r_{\mathbf x}$ for a new dataset by 
computing its similarity to the training datasets in the $\mathcal X$ feature space, similarity that was learned in a
manner that reflects the datasets similarity in terms of their relative workflow preferences.


\subsection{Learning a data mining workflow metric}
\label{sec:wsim}
To learn a Mahalanobis metric matrix $\mathbf W$ in the $\mathcal A$ data mining workflow
space we will proceed in exactly the same manner as we did with the datasets 
using now the $\mathbf R^T\mathbf R$ matrix the elements of which will give the rank correlation 
coefficients of the dataset preference vectors of the workflows, measuring thus the 
similarity of workflows in terms of their relative performance over the different datasets. 
More precisely as before we start with the metric learning optimization problem:
\begin{eqnarray*}
\label{eq:metric-wf.1}
\min_{{\mathbf W}_{\mathcal A}} F_2(\mathbf W_{\mathcal A})
     & = & ||\mathbf{R}^\text{T} \mathbf{R} - \mathbf{A} \mathbf{W}_{\mathcal A} \mathbf{A}^\text{T}||_F^2 + \mu_1 tr(\mathbf{W}_{\mathcal A}) \\ \nonumber
s.t. &   & \mathbf W_{\mathcal A} \succeq 0 \nonumber
\end{eqnarray*} 
which we cast to the problem of learning a linear transformation $\mathbf V$ in the workflow space as:
\begin{eqnarray}
\label{eq:metric-wf.2}
\min_{\mathbf V} F_2(\mathbf V)
     & = & ||\mathbf{R}^\text{T} \mathbf{R} - \mathbf{A} \mathbf{V} \mathbf V^T \mathbf{A}^\text{T}||_F^2 + \mu_1 ||\mathbf{V}||_F^2 
\end{eqnarray} 
where $\mathbf W_{\mathcal A}= \mathbf V \mathbf V^T$ is the $l \times l$ 
metric matrix, and $\mathbf V$ an associated linear transformation with 
dimensionality $l \times t$ that projects workflow descriptions into a 
new space of $t$ dimensionality.  As before this is 
not a convex optimization problem. We will solve it using gradient descent.
Similar to the dataset case using the learned metric the similarity of two  workflows $\mathbf a_i$ and $\mathbf a_j$ is  
$\mysim_{\mathcal A}(\mathbf a_i, \mathbf a_j) = \mathbf a_i \mathbf V \mathbf V^T \mathbf a_j$. Given some
new workflow $\mathbf a$ its workflow neighborhood $N_{\mathbf a}$ consists of 
the $N$ workflows that are most similar to $\mathbf a$ with respect to the similarity of their 
relative dataset preferences as this is computed in the original feature space $\mathcal A$. 
With the help of $N_{\mathbf a}$ we can now compute the dataset preference vector of $\mathbf a$
as the weighted average of the dataset preference vectors of its nearest neighbors by: 
\begin{eqnarray}
\mathbf r_{\mathbf a} = \zeta_{\mathbf a}^{-1} \sum_{a_i \in N_{a}} \mathbf r_{\mathbf a_i} \mysim_{\mathcal A}(\mathbf a,\mathbf a_i)
\end{eqnarray}
where $\zeta_{\mathbf a}$ is a normalization factor given by $\zeta_{\mathbf A} = \sum_{\mathbf A_i \in A} \mysim_{\mathcal A}(\mathbf a, \mathbf a_i)$.
Thus using the learned metric we can compute the dataset preference vector $\mathbf r_{\mathbf a}$ for a new workflow by 
computing its similarity to the training workflow in the $\mathcal A$ feature space, similarity that was learned in a
manner that reflects the workflows similarity in terms of their relative dataset preferences.

\subsection{Learning a heterogeneous metric over datasets and workflows}
\label{sec:dwsim}
The last metric that we want to learn is one that will relate datasets to data mining workflows reflecting the 
appropriateness/preference of a given workflow for a given dataset in terms of the relative performance of the
former applied to the latter. We will do so by starting with the following metric learning 
optimization problem 
\begin{eqnarray*}
\label{eq:metric-ds-wf.1}
\min_{{\mathbf W}} F_3(\mathbf W)
     & = & ||\mathbf{R} - \mathbf{X} \mathbf{W} \mathbf{A}^\text{T}||_F^2 + \mu_1 tr(\mathbf{W}) \\ \nonumber
s.t. &   & \mathbf W \succeq 0 \nonumber
\end{eqnarray*}
which if we parametrize the $d \times l$ metric matrix $\mathbf W$ with the help of two linear transformation
matrices $\mathbf U$ and $\mathbf V$ with dimensions $d\times t$ and $l \times t$ can be rewritten as:
\begin{eqnarray}
\label{eq:metric-ds-wf.2}
\min_{{\mathbf{U,V}}} F_3(\mathbf {U, V})
     & = & ||\mathbf{R} - \mathbf{X} \mathbf{U V}^T \mathbf{A}^\text{T}||_F^2 + \\ \nonumber 
     &   & \mu_1 ||\mathbf{U}||_F^2 + \mu_2||\mathbf{V}||_F^2 \nonumber
\end{eqnarray}
Essentially what we do here is to project the descriptions of datasets and workflows to a common space with 
dimensionality $t$ over which we compute their similarity in a manner that reflects the preference matrix
$\mathbf R$. We will set $t$ to the $\min(rank(\mathbf A), rank(\mathbf X))$.
In other words we learn a heterogeneous metric which computes similarities of datasets and 
workflows in terms of the relative performance of the latter when applied on the former.
\note[Alexandros]{I think that the nature of the matrix $\mathbf R$ as is now is problematic. There is 
now way to compare the values of different pairs of algorithms, datasets, between them in order to get 
a total order. What I mean is that $R(\mathbf x_j, \mathbf a_i)$ is not comparable to $R(\mathbf x'_j, \mathbf a'_i)$
Maybe we need a way to reformalize, renormalize $\mathbf R$ in a way that it gives a total
order between different pairs.}
Using the new similarity metric we can now compute directly the match between a dataset $\mathbf x$ and a workflow $\mathbf a$
as:
\begin{eqnarray}
r_{\mathbf x, \mathbf a} = \mathbf x \mathbf U \mathbf V^T \mathbf a
\end{eqnarray}
Clearly we can use this not only to determine the goodness of match between a dataset and a data mining workflow
but also given some dataset and a set of workflows to order the latter according to their appropriateness with 
respect to the former, thus solving the meta-mining task 1, and vice versa given a workflow and a set of datasets to order the
latter according to their appropriateness for the former thus solving meta-mining task 2.

In the objective function of the optimization problem~(\ref{eq:metric-ds-wf.2}) we focus exclusively on trying to 
learn a metric that will reflect the appropriateness of some workflow for some dataset as this is given by the 
entries of the $\mathbf R$ preference matrix. However there is additional information that we can bring in if
we exploit the objective functions of the optimization problems~(\ref{eq:metric-dataset.2}) 
and~(\ref{eq:metric-wf.2}) and use them to additionally regularize the objective function of (\ref{eq:metric-ds-wf.2}). 
The overall idea here is that we will learn three different metrics in the spaces
of datasets, workflows, and datasets-workflows, all of them parametrized by two linear transformations in a 
manner that will reflect the basic meta-mining assumptions, namely that similar datasets should have similar
workflow preference vectors, similar workflows should have similar dataset preference vectors, and that the
heterogeneous metric between datasets and workflows should reflect the appropriateness of datasets for workflows.
By combining the three optimization problems of~(\ref{eq:metric-dataset.2}),~(\ref{eq:metric-wf.2}), and ~(\ref{eq:metric-ds-wf.2})
we get the following metric learning optimization problem that achieves these goals:
\begin{eqnarray}
\label{eq:metric-ds-wf-ext}
\min_{{\mathbf{U,V}}} F_4(\mathbf {U, V}) & = & \alpha F_1(\mathbf {U}) +  \beta F_2(\mathbf {V}) + \gamma F_3(\mathbf {V}, \mathbf {U}) \\ \nonumber
     & = & \alpha ||\mathbf{R} \mathbf{R}^\text{T} - \mathbf{X} \mathbf{U} \mathbf U^T \mathbf{X}^\text{T}||_F^2  \\ \nonumber
     & + & \beta ||\mathbf{R}^\text{T} \mathbf{R} - \mathbf{A} \mathbf{V} \mathbf V^T \mathbf{A}^\text{T}||_F^2  \\ \nonumber
     & + & \gamma ||\mathbf{R} - \mathbf{X} \mathbf{U V}^T \mathbf{A}^\text{T}||_F^2   \\ \nonumber 
     & + & \mu_1 ||\mathbf{U}||_F^2 + \mu_2||\mathbf{V}||_F^2 \nonumber
\end{eqnarray}
where $\alpha, \beta, \gamma$, are positive parameters that control the importance of the three different optimization terms.
As it was the case with optimization problem~(\ref{eq:metric-ds-wf.2}) this optimization problem can also be used to address
all three meta-mining tasks. In fact (\ref{eq:metric-ds-wf-ext}) is the most general formulation of the metric-learning
based hybrid reccomendation problem and includes as special cases problems~(\ref{eq:metric-dataset.2}) and~(\ref{eq:metric-wf.2}).

Matrix factorization, often used in recommender systems, also learns a decomposition 
of a matrix to component matrices $\mathbf U$ and $\mathbf V$ under different constraints.
However, by its very nature it cannot handle well the out-of-sample problem.  
The objective function of  problem (\ref{eq:metric-ds-wf-ext}) uses
as additional constraints the objective functions of (\ref{eq:metric-dataset.2}) and~(\ref{eq:metric-wf.2})
and learns a common space for the datasets and workflows, which are induced by the projection 
matrices $\mathbf U $ and $\mathbf V$. As a result, the out-of-sample problem,
i.e. cold start problem in recommender system, is naturally
handled by the optimization problem~(\ref{eq:metric-ds-wf-ext}).

\section{Dataset and workflow descriptors}
\label{sec:metadata}

In the following two sections we will describe the dataset and workflow descriptors that we will use in our meta-mining experiments.

\subsection{Dataset Descriptors}

Originally proposed by the STATLOG project~\cite{King1995}, the idea of characterizing datasets has been the main stream in meta-learning during these 
last decades~\cite{Soares2000,Kopf2000,Hilario2001,Kalousis2004}. Various characterizations have been subsequently proposed, from which we have selected 
the most relevant ones summarized as follows: \\
{\em statistical measures}: number of instances, number of classes, proportion of missing values, proportion of continuous / categorical features, noise signal ratio. \\
{\em information-theoretic measures}: class entropy, mutual information. \\
{\em geometrical and topological measures}~\cite{HoBasu2006}: non-linearity, volume of overlap region, maximum fisher's discriminant ratio, 
fraction of instance on class boundary, ratio of average intra/inter class nearest neighbour distance. \\
{\em model-based measures}: error rates and pairwise $1-p$ values obtained by landmarkers~\cite{Pfahringer2000} such as ZeroR, one-nearest-neighbor, 
Naive Bayes, Decision Stumps~\cite{Iba1992}, Random Trees~\cite{Breiman2001}, and the linear SVM~\cite{Cortes1995}, and the distributions of the weights learned by the Relief~\cite{Robnik-Sikonja2003} and SVMRFE~\cite{Guyon_Weston_Barnhill_Vapnik_2002} feature selection algorithms. \\
\note[Alexandros]{Give the full list of landmarkers that you used. Also given references for the different algorithms, especially Relief and SVM-RFE.}
Overall, we use a large spectrum of dataset characteristics, from very simple ones such as the number of instances to more elaborated ones such as 
the model-based measures, giving a total number of $d=150$ dataset characteristics.

\subsection{Workflow descriptors}
The ability to describe data mining algorithms and workflows and use these descriptors for meta-learning and meta-mining is a very recent development~\cite{Hilario2010}.
There the authors used DMOP, a data mining ontology, to describe learning algorithms and data-processing algorithms such as feature selection, discretization and normalization, 
with respect to the mathematical concepts they implement and different properties, such as their bias/variance profile, their sensitivity to the type of attributes, their learning 
strategy, etc.  In addition the same ontology allows to anotate operators (algorithm implementations) of data mining workflows with their respective concepts. 
A data mining workflow is typically a direct acyclic graph of data mining operators.

\begin{figure}
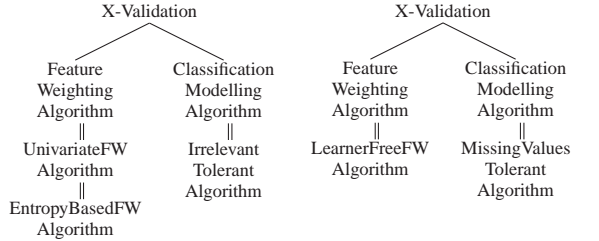
 
\centering
\setlength{\fboxrule}{0.1pt}
\label{fig:patt1}
\let\qtreeinithook=\scriptsize
\Tree [ [ [ {EntropyBasedFW\\Algorithm} !{\brSubsume} ].{UnivariateFW\\Algorithm} ].{Feature\\Weighting\\Algorithm} !{\brRestore} [ {Irrelevant\\Tolerant\\Algorithm} !{\brSubsume} ].{Classification\\Modelling\\Algorithm} !{\brRestore}   ].X-Validation
\label{fig:patt2}
\let\qtreeinithook=\scriptsize
\Tree [ [ {LearnerFreeFW\\Algorithm} !{\brSubsume} ].{Feature\\Weighting\\Algorithm} !{\brRestore} [ {MissingValues\\Tolerant\\Algorithm} !{\brSubsume} ].{Classification\\Modelling\\Algorithm} !{\brRestore}  ].X-Validation
\caption{Two workflow patterns with cross-level concepts. Thin edges depict workflow decomposition; double lines depict DMOP's concept subsumption.}
\label{fig:pattern}
\end{figure}

In order to describe the data mining workflows we follow the propositionalization approach used in~\cite{Hilario2010}. We derive from the annotated direct acyclic graphs that describe
the data mining workflows a set of frequent closed workflow patterns using the tree-structured apriori algorithm of~\cite{Zaki05}. The description of a workflow is then given by a 
binary vector that indicates the presence or absence of each of the frequent patterns; the final workflow description contains $l=214$ features. In figure~\ref{fig:pattern} we give two 
examples of workflow patterns that have been abstracted from ground feature selection + classification workflows based on DMOP’s algorithm hierarchy. These patterns help us understand how the workflow space 
is structured by describing frequent workflow structures using the DMOP concepts.

\section{Experiments}
\label{sec:exp}

\note[Removed]{
\begin{table*}
\centering
\begin{tabular}{|l|ll|ll|ll|} 
\multicolumn{7}{c}{Task 1.} \\ \hline
  & $\rho$ & $\diff$  & $\acc$ & $\diff$ & $\mae$ & $\diff$ \\ \hline
def & 0.332 & -  & 77.8  &  - & 4.50 & - \\ \hline 
$F_1$ & {\bf 0.366} & 34/65 & {\bf 78.6}  & 33/65 & 4.83  & 40/65 \\ 
 & & p=0.804 & & p=1 & & p=0.082 \\ \hline
$F_3$ & 0.286 & 23/65 & 77.1  & 23/65 & 5.64 & 19/65 \\ 
 & & p=0.975 & & p=0.975 & & p=0.999 \\ \hline
$F_4$ & {\bf 0.391} & 40/65 & {\bf 79.3}  & 41/65 & {\bf 4.22} & 46/65 \\ 
 & & p=0.082 & & {\bf p=0.047} & & {\bf p=0.001} \\ \hline
\end{tabular}
\begin{tabular}{|l|ll|ll|ll|} 
\multicolumn{7}{c}{Task 2.} \\ \hline
  & $\rho$ & $\diff$ & $\acc$ & $\diff$ & $\mae$ & $\diff$  \\ \hline
def & NA & -  &  NA & - & 4.84 & -  \\ \hline 
$F_2$ & 0.445 & - & 83.0 & - & {\bf 3.90}  & 29/35 \\ 
& & - & & - & & {\bf p=1e-3} \\ \hline
$F_3$ & 0.478 & - & 82.6 & - & {\bf 4.47} & 26/35 \\ 
& & - & & - & & {\bf p=0.006} \\ \hline
$F_4$ & 0.491 & - & 82.4 & - & {\bf 4.39} & 28/35 \\ 
& & - & & - & & {\bf p=0.005} \\ \hline
\end{tabular}
\begin{tabular}{|l|ll|} 
\multicolumn{3}{c}{Task 3.} \\ \hline
   & $\mae$ & $\diff$ \\ \hline
{\em def} & 4.84 & -  \\ \hline 
& & \\
& & \\ \hline
$F_3$ & 6.46 & 880/2275 \\ 
& & p=1 \\ \hline
$F_4$ & 4.99 & 1070/2275 \\ 
& & p=0.999 \\ \hline
\end{tabular}
\caption{Evaluation of the three meta-mining tasks on the 65 biological datasets.}
\label{table:bio}
\end{table*} }

\begin{table*}
\centering
\begin{tabular}{|c|c|c|c|c|c|c|c|c|c|} \hline
1 & 2 & 3 & 4 & 5 & 6 & 7 & 8 & 9 & 10 \\ \hline
{\em LR} & {\em IG}+{\em LR} & {\em RF}+{\em LR} & {\em SVMRFE}+{\em SVM$_l$} & {\em SVMRFE}+{\em LR} & {\em IG}+{\em NBN} & {\em IG}+{\em SVM$_l$} & {\em CHI}+{\em NBN} & {\em SVMRFE}+{\em NBN} & {\em RF}+{\em SVM$_l$}  \\ \hline
 25 & 12 & 13  & 16  & 13  & 17  & 14  & 10  & 13  & 8 \\ \hline
\end{tabular}
\caption{Default top-10 workflows and their frequency in the top-5 positions.}
\label{table:top5}
\end{table*}

\begin{table*}
\centering
\begin{tabular}{|l|l|l|l|} 
\multicolumn{4}{c}{Learning Workflow preferences} \\ \hline
      		& $\rho$      & $\acc$          & $\mae$      \\ \hline
{\em def}   & 0.332       & 77.8       		& 4.50        \\ \hline 
{\em EC} & {\bf 0.356}    & 77.8 			& {\bf 4.39}   \\
$\diff$	 & 32/65 p=1	  & 32/65 p=1	    & 37/65 p=0.321 \\ \hline
$F_1$ 	 & {\bf 0.366}    & {\bf 78.6}      & 4.83  \\ 
$\diff$  & 34/65 p=0.804 & 33/65 p=1        & 40/65 p=0.082 \\ 
$\diff_{EC}$ & 35/65 p=0.620 & 	33/65 p=1	& 20/65 p=0.003 \\ \hline
$F_3$ 	 & 0.286          & 77.1            & 5.64  \\ 
$\diff$  & 23/65 p=0.025  & 23/65 p=0.025   & 19/65 p=0.001 \\ 
$\diff_{EC}$ & 19/65 p=0.001 & 	27/65 p=0.215	& 14/65 p=1e-6	 \\ \hline
$F_4$ 	 & {\bf 0.391}    & {\bf 79.3}      & {\bf 4.22}  \\ 
$\diff$  &  40/65 p=0.082 & 41/65 {\bf p=0.047} & 46/65 {\bf p=0.001} \\ 
$\diff_{EC}$ & 42/65 {\bf p=0.025}   & 44/65 {\bf p=0.006} & 42/65 {\bf p=0.025}	 \\ \hline
\end{tabular}
\quad
\begin{tabular}{|l|l|l|} 
\multicolumn{3}{c}{Learning Dataset preferences} \\ \hline
       		& $\rho$ 			&          $\mae$    \\ \hline
{\em def}   & NA   				&          4.84         \\ \hline 
{\em EC}  	& 0.375    			&          {\bf 4.29}   \\ 
$\diff$     & NA                & 	30/35 {\bf p=2e-5} \\ \hline
$F_2$  		& 0.445    			&          {\bf 3.90}   \\ 
$\diff$     & NA                & 	29/35 {\bf p=1e-3} \\ 
$\diff_{EC}$& 23/35 p=0.09      & 	22/35 p=0.176 \\ \hline
$F_3$  		& 0.478        		&          {\bf 4.47}   \\ 
$\diff$     & NA                & 26/35 {\bf p=0.006} \\ 
$\diff_{EC}$& 22/35 p=0.176     & 17/35 p=1 \\ \hline
$F_4$  		& 0.491       		&          {\bf 4.39}  \\ 
$\diff$     & NA                & 28/35 {\bf p=0.005} \\ 
$\diff_{EC}$& 24/35 {\bf p=0.041}  & 18/35 p=1 \\ \hline
\end{tabular}
\quad
\begin{tabular}{|l|l|} 
\multicolumn{2}{c}{Learning DS-WF preferences} \\ \hline
   & $\mae$ \\ \hline
{\em def} & 4.84    \\ \hline 
- & - \\
- & - \\ \hline
- & - \\ 
- & - \\ 
- & - \\ \hline
$F_3$ & 6.46 \\ 
$\diff$ & 880/2275 p=0 \\ 
- & - \\ \hline
$F_4$ & 4.99  \\ 
$\diff$ & 1070/2275 p=0.005 \\ 
- & - \\ \hline
\end{tabular}
\caption{Evaluation results. $\delta$ and $\delta_{EC}$ denote comparison results with the default ({\em def}) and the Euclidean 
baseline strategy ({\em EC}) respectively. $\rho$ is the Spearman's rank correlation coefficient, in t5p we give 
the average accuracy of the top five workflows proposed by each strategy, and mae is the mean average error. X/Y indicates the 
number of times X that a method was better overall the experiments Y than the default or the baseline
strategy.}
\label{table:bio}
\end{table*}


In this section, we will perform a systematic evaluation to examine the performance of the different 
metric learning optimization problems for meta-mining that we presented in the previous sections. More
precisely we will evaluate the performance of the dataset metric learning optimization problem given in~(\ref{eq:metric-dataset.2})
to the meta-mining task of learning workflow preferences for a given dataset; the performance of 
the workflow metric learning optimization problem of (\ref{eq:metric-wf.2}) to the meta-mining task of learning dataset
preferences; and finally the performance of the two metric learning optimization problems, (\ref{eq:metric-ds-wf.2}),~(\ref{eq:metric-ds-wf-ext}), 
for all three meta-mining tasks.

\subsection{Base-level Experiments}

In order to meta-mine we first need to perform a set of base-level experiments over which we will construct our meta-mining models. To do so we used 
65 real world cancer microarray datasets, most of them were taken from the National Center for Biotechnology Information~\footnote{http://www.ncbi.nlm.nih.gov/}. \note[Alexandros]{Add a reference, on where these data were obtained from.} 
Microarray datasets are characterized by a high-dimensionality and a small sample size, and a relatively low number of classes, most often two. 
These datasets have an average of 79.26 instances, 15268.57 attributes, and 2.33 classes.
\note[Alexandros]{Add a sentence given the average number of instances, attributes, and classes per dataset.} 
On these datasets we applied a total of 35 classification data mining workflows; 
28 of them were workflows that contained one feature selection and one classification algorithm, while the seven remaining ones had only a single 
classification algorithm. We used the four following feature selection algorithms: 
Information Gain, {\em IG}, 
Chi-square, {\em CHI}, 
ReliefF~\cite{Robnik-Sikonja2003}, {\em RF}, 
and recursive feature elimination with SVM~\cite{Guyon_Weston_Barnhill_Vapnik_2002}, {\em SVMRFE}, 
and fixed the number of selected features to ten.  For classification we used the seven following algorithms: 
one-nearest-neighbor,  {\em 1NN}, 
the {\em C4.5}~\cite{Quinlan93} and {\em CART}~\cite{Breiman84} decision tree algorithms, 
a Naive Bayes algorithm with normal probability estimation, {\em NBN}, 
a logistic regression algorithm, {\em LR}, 
and SVM~\cite{Cortes1995} with the linear, {\em SVM$_l$} 
and the rbf, {\em SVM$_r$}, 
kernels.  We used the implementations of these algorithms provided by the RapidMiner data mining suite with their default parameters.  
Overall we had a total of $65 \times (28+7) = 2275$ base-level DM experiments, i.e. applications of these workflows on the datasets.
To construct the $\mathbf R$ preference matrix we estimated the performance of the workflows using 10-fold cross-validation and applied 
the scoring McNemar based scoring schema described in section~\ref{sec:task}.
In table~\ref{table:top5} we give for each of the ten top workflows over the full set of 65 datasets 
the number of times that these were ranked in the top five positions.


\subsection{Baseline Strategies and Evaluation Methodologies}

In order to assess how well the different variants perform we need to compare them with some 
default and baseline strategies. \note[Alexandros]{Ideally we should also compare them with another meta-learning/meta-mining method
which uses a simple modelling, such as the one we did for the book. But now it is too late.}
For the meta-mining task of workflow preference learning, we will use as the default strategy
the preference vector given by average of the workflow preference vectors over the 
different training datasets for a given testing dataset. We should note that this is
a rather difficult baseline to beat since the different workflows will be ranked 
according to their average performance on the training datasets, with workflows that
perform consistently well ranked on the top.  
\note[Alexandros]{It might be interesting 
to give some statistics on the top performing workflows, e.g. a table that gives the five 
or ten best and how many times they were ranked on the top position.}
For the second task of providing a dataset preference vector for a given testing workflow we have
a similar default strategy, we will use the average of the dataset preference vectors over the different 
training workflows.
However this strategy for the workflows leads to a trivial constant vector of dataset preferences
due to the fact that the total sum of workflow points for a given dataset is fixed to $m(m-1)/2$, 
when we compare $m$ workflows, by the very same nature of the workflow ranking schema for a given 
dataset.
Finally for the last meta-mining task we will use as the default strategy for the prediction for the appropriateness 
of a workflow for a dataset the average over the values of the preference matrix of the training set. 
We will denote the default strategy used in the three meta-mining tasks by {\em def}. In 
addition we will also have as a baseline strategy the provision of recommendation when we
use a simple Euclidean distance, i.e. all attributes are treated equally and there is no 
learning, which we will denote by {\em EC}. However this baseline is only applicable to the 
first two meta-mining tasks, learning workflow preferences and learning dataset preferences, since it cannot be applied to the
kind of heterogeneous similarity problem that we have in the third meta-mining task.

As resampling techniques we will use leave-one-dataset-out to estimate the performance on 
the workflow preference learning task, leave-one-workflow-out for the dataset preference learning
task, and leave-one-dataset-and-one-workflow-out for the third task of predicting the appropriateness
of a workflow for a dataset.

\note[Removed]{
our meta-mining approaches perform, we need to compare them with some baseline.
To define the baseline for tasks 1 and 2, we will use the average scores of the workflows/datasets over all past DM experiments:  
for task 1, we will average all dataset score vectors $\mathbf{r}_x \in \mathbf{R}$ that we will denote by $\tilde{\mathbf{r}}_{x_n}$, 
in order to define the default score vector for a given new dataset $x_n$. 
This default score vector will deliver as a majority rule the best performing workflows over all past datasets and   
thus, it should be regarded as a high-level baseline since it might provide by default high performance accuracies on the dataset to be evaluated. 
For task 2, we will average all workflow score vectors $\mathbf{r}_a^\text{T} \in \mathbf{R}$ for a given new workflow $a_n$ that we will denote 
by $\tilde{\mathbf{r}}_{a_n}^\text{T}$.  Note that this default score vector for workflows can lead to a constant vector $\mathbf{L}$ if there are 
no missing scores in the score matrix $\mathbf{R}$, otherwise it will point to the datasets having 
the maximum of average score.  Finally, for task 3, we will define the default scalar score for a new pair of dataset-workflow $(x_n,a_n)$ as the 
average of all scores in the score matrix $\mathbf{R}$  that we will denote by $\mu_{\mathbf{R}}$.  

To evaluate our approaches, we will do for task 1 a leave-one-dataset-out, for task 2 a leave-one-workflow-out, and for task 3 a leave-one-dataset+workflow-out, 
where we will compare the respective baseline strategies with the score predictions of our approaches for each task on the left-out entity,  
using the following evaluation measures. 
}

To quantify the performance we will use a number of evaluation measures. For the first two meta-mining tasks
we will report the average Spearman's rank correlation coefficient between the predicted preference vector and 
the real preference vector over the testing instances. We will denote this average by $\rho$.
This measure will indicate the degree to which the different methods predict correctly the preference 
order. Note that this quantity is not computable for the default strategy in the case of the learning 
dataset preferences task, due to the fact that the dataset preference vector that it produces is fixed,
as we explained previously, and the Spearman rank correlation coefficient is not computable when one of
the two vectors is fixed. In addition to the Spearman rank correlaction coefficient 
for the meta-mining task of learning workflow preferences we will also report the 
average accuracy of the top five workflows suggested by each method, measure which we will denote by {\em t5p}.
Finally for the three meta-mining tasks we will also report the mean average error, mae, over the respective testing
instances, of the predicted values for $\mathbf r_{\mathbf x}$, $\mathbf r_{\mathbf a}$, and $\mathbf r_{\mathbf x, \mathbf a}$,
for learning workflow preferences, dataset preferences, and dataset-workflow preferences, respectively, and the true values.
For each measure, method, and meta-mining task, we will give the number of times that the method was better than the respective
default and baseline strategies over the total number of datasets, workflows, or dataset, workflow pairs (depending on the meta-mining task),
as well as the statistical significance of the result under a binomial test with a statistical significance level of 0.05. The 
comparison results with the default strategy will be denoted by $\delta$ while the comparison to the Euclidean baseline by $\delta_{EC}$.

\subsection{Experiment Results on the Biological Datasets}
We will now take a close look on the experimental results for the different meta-mining tasks 
and objective functions that we have presented to address them. The full results are given in
Table \ref{table:bio}. 

\paragraph{Learning Workflow Preferences}
Learning algorithm preferences is the most popular formulation in the traditional stream of meta-learning. There
given a dataset description we seek to identify the algorithm that will most probably deliver the best results for
the given dataset. In that sense this meta-mining task is the most similar to the typical meta-learning task. We have presented
three different objective functions that can be used to address this problem. $F_1$, optimization problem~(\ref{eq:metric-dataset.2}), 
makes use of only the dataset descriptors and learns a similarity measure in that space that best approximates
their similarity with respect to their relative workflow preference vectors. In traditional meta-learning this similarity
is computed directly in the dataset space, it is not learned, and most importantly it does not try to model the relative 
workflow preference vector, \cite{Soares2000,Kalousis99}. In our experimental setting the strategy that implements this 
traditional meta-learning approach is the Euclidean distance-based dataset similarity, {\em EC}. In addition to the homogeneous metric learning 
approach we can also use the two heterogeneous metric learning variants to provide the workflow preferences. The simplest
one, corresponding to the optimization function $F_3$, optimization problem~\ref{eq:metric-ds-wf.2}, uses both dataset 
and workflow characteristics and tries to directly approximate the relative preference matrix. However this approach
ignores the fact that the learned metric should reflect two basic meta-mining requirements, that similar datasets should have 
similar workflow preferences, and that similar workflows should have similar dataset preferences. The optimization
function $F_4$, optimization problem~\ref{eq:metric-ds-wf-ext}, reflects exactly this bias by regularizing appropriately 
the learned metrics in the dataset and workflow spaces so that they reflect well the similarities of the respective preference 
vectors. Before discussing the actual results, given in the left table of Table~\ref{table:bio}, we give the parameter settings for
the different variants. $F_1$: $\mu_1 = 0.5$, $N_{x_n} = 5$; $F_3$: $\mu_1=\mu_2=0.5$; $F_4$: $\alpha=1e^{-10}$, 
$\beta=1e^{-3}$, $\gamma=1e^{-3}$, $\mu_1 = 10$, $\mu_2 = 0$.
\note[Alexandros]{This parameter setting is rather problematic, and I think we will have a problem with the reviewers here. 
It would have been more appropriate to try to cross-validate some of them. In addition the different components should be
on the same scale, otherwise there is no sense in trying to interpret the weightings.}
These parameters reflect what we think are appropriate choices based on our prior knowledge 
of the meta-mining problem. Better results would have been obtained if we had tuned, at least 
some of them, via inner cross validation.

\note[Removed]{
From the standard meta-learning perspective, algorithm/workflow selection involves only as meta-knowledge the use of dataset descriptors. 
Meta-mining perspective makes additionally use of workflow descriptors to enhance the learning of workflow preferences.
In this section, we will make a systematic comparison of these two perspectives. 
Specifically, we will compare the results of the methods focusing only on dataset descriptors, 
given by the objective function $F_1$ and the {\em EC} metric, versus the results of methods 
combining both descriptors in a heterogenous manner, the $F_3$ and $F_4$ functions.
We will do the same for the workflow descriptors in the next section.
We set the parameters of the above objective functions as follows.  
$F_1$: $\mu_1 = 0.5$, $N_{x_n} = 5$, $F_3$: $\mu_1 = 0.5$, $F_4$: $\alpha=1e^{-10}$, $\beta=1e^{-3}$, $\gamma=1e^{-3}$, $\mu_1 = 10$,  $\mu_2 = 0$.
The parameter setting for the $F_4$ function reflects the prior knowledge we had on the meta-features. 
In particular, we noted that the bio-datasets descriptors had small variance because of the similar nature underlying these learning problems. 
Thus to learn a fully heterogenous metric in the objective function $F_4$, 
we gave more importance on the workflow descriptors than on the dataset descriptors with the $\beta$ and $\alpha$ parameters  
and with the regularization parameters $\mu_1$ and $\mu_2$. 
}

Looking at the actual results we see right away that the approach that makes use of only the dataset characteristics, $F_1$, 
has a performance that is not statistically significant different neither from the default, nor from the $EC$ baseline with 
respect to the Spearman's rank correlation coefficient, $\rho$, and the average accuracy of the top five workflows it 
suggests, tp5. In addition it is statistically significant worse than the $EC$ with respect to the mean average error criterion, mae, 
having a lower mae value than {\em EC} only in 20 out of the 65 datasets. Looking at the performance of the heterogeneous
metric that tries to directly approximate the preference matrix $\mathbf R$, we see that its results are 
quite disappointing. It is significant worse than the default strategy and the {\em EC} baseline for almost all performance 
measures. So trying to learn a heterogeneous metric that relies exclusively on the approximation of the preference matrix is 
definitely not an option. However when we turn to the $F_4$ objective function that learns the heterogeneous metrics in 
a manner that they do not only reflect the preference manner, but also the fact that similar datasets should have similar
workflow preferences and vice versa, there we see that the performance we get is excellent. $F_4$ beats in a statistically 
significant manner both the default strategy as well as the {\em EC} baseline in almost cases, the only exception is the 
Spearman's correlation coefficient comparison with the default where the level of significance is high, $p=0.082$, but does
not overpass the significance threshold of 0.05. Overall in such a recommendation scenario the best strategy consists in 
learning a combination of the two homogeneous and one heterogeneous metrics that reflect the similarities of the 
datasets with respect to the workflow preferences, the similarities of the workflows with respect to the dataset 
preference vectors, as well as the similarities of workflows-datasets according to the preference matrix.

\note[Removed]{
The results of task 1 are given in the left table of Table \ref{table:bio}. 
First, we see that $F_4$ gives  the best results for this task.  
We have significant improvement on the three evaluation measures compared with the baseline (i.e. the best performing workflows in our collection of datasets) and also with {\em EC}, $F_1$ and $F_3$.  
When we compare the results of $F_4$ with those of {\em EC} and $F_1$, we see that the former approach improve the evaluation measures in a significant manner over the two latters. 
When we compare $F_4$ with $F_3$, we see as well strong differences in the evaluation measures in favor of the former. 
The latter has poor results as it is not regularized in its fitness of past preferences by dataset and workflow similarities.  
Thus  heterogenous metric as given by $F_4$ greatly improves homogenous dataset metrics, as well as of $F_3$, and provides good conclusions for this meta-mining approach.

In contrast, the homogenous metrics, {\em EC} and $F_1$,  which are comparable in their results give together another picture. 
Compared with the baseline strategy, we have higher $\rho$ and $\acc$ evaluation measures but without significant results.
Specifically, the $\diff$ measure shows that these improvements occur only on one half of the bio-datasets.   
The $F_1$ and {\em EC} methods, if able to determine the workflow preferences on half of the datasets according to the dataset descriptors, do not do better than the baseline on the other half.  
Thus these two dataset homogenous metrics are not reliable enough for learning workflow preferences in the case of our bio-dataset collection.



}

\paragraph{Learning Dataset Preferences}
The goal of this meta-mining task is given a new workflow and a collection of 
datasets to provide a dataset preference vector that will reflect the order of appropriateness of the datasets for the given workflow.
As already mentioned the default strategy provides here a vector of equal ranks thus we cannot compute its Sperman's rank correlation
coefficient. 
\note[Alexandros]{One comment that I have here is that the actual computation of the similarities of the dataset preference
vector seems to be problematic, since I am not sure that these vectors are really comparable across different workflows.
The same comment is also valid when we try to compare the $r_{\mathbf a, \mathbf x}$ values over different datasets. We 
need a good way to do this kind of normalization, that will make the comparison of values meaningful.}
We will compare the performance of the $F_2$ objective function that makes use of only of the workflow descriptors when 
it tries to approximate the similarity of the dataset preference vectors, and these of $F_3$ and $F_4$. 
We used the following parameter: $F_2$: $\mu_1 = 10$, $N_{a_n} = 5$; $F_3$: $\mu_1 =\mu_2 = 10$;
$F_4$: $\alpha=1e^{-10}$, $\beta=1e^{-3}$, $\gamma=1e^{-3}$, $\mu_1 = 0.5$,  $\mu_2 = 0$.
Looking at the results, middle table of Table~\ref{table:bio}, we see that when it comes to the mean average error, 
all methods achieve a performance that is statistically significant better than that of the default strategy, suggesting
that this meta-mining task is probably easier than the first one. This makes sense since it is easier to describe 
a workflow similarity in terms of the concepts that these workflows use, than what it is to describe a dataset similarity
in terms of the datasets characteristics. Neither $F_3$ nor $F_4$ have a mae performance that is statistically significant 
better than the Euclidean baseline. Nevertheless $F_4$ is statistically significant better than the Euclidean when it comes
to the Sperman's rank correlation coefficient. Thus for this meta-mining task there is also evidence that we should take 
a more global approach by accounting for all the different constraints on the dataset and workflow metrics as $F_4$ does.

\note[Removed]{
For the second meta-mining task, we will follow the same systematic comparison between homogenous and heterogenous metrics.  
Our prior was to put more importance on the dataset descriptors than in the previous task in order to see if these descriptors 
will help for addressing the task of learning dataset preferences.  We did that by regularizing more the workflow patterns, 
while at the same time decreasing the regularizer on the dataset descriptors.

The results for task 2  give now another picture compared to those of task 1.  These are given in the middle table of Table \ref{table:bio}.
Compared with the baseline, we get a significant improvement on the $\mae$ evaluation measure for all of our three objective functions 
$F_2$, $F_3$ and $F_4$ and the Euclidean metric, {\em EC}. These results suggest that we have a much more structured and coherent space 
within the workflow descriptors than within the dataset descriptors. The well-defined workflow descriptors allow to 
retrieve easily for a given workflow, say one containing the {\em C4.5} decision tree algorithm, the datasets on which a similar workflow, 
say one with a {\em CART} algorithm, has good performance. Therefore we get with the homogenous metrics, $F_2$ and {\em EC}, the lowest 
$\mae$ 
because we learnt with them an optimal workflow similarity measure. 

In contrast, the heterogenous metrics, $F_3$ and $F_4$, have both an higher $\mae$ but at the same time they also 
achieve higher correlations than the homogenous measures.  These results 
might be understood as the combination of workflow descriptors and dataset descriptors which induce a better order on datasets 
than the use of workflow descriptors only, but does not achieve higher prediction of the preference values themselves. 
Nevertheless, $F_4$ achieves the highest correlation as it is regularized by dataset and workflow similarities.
}

\paragraph{Learning Dataset-Workflow Preferences}
The last meta-mining task is by far the most difficult one. Here we want to predict the appropriateness of a new workflow 
for a new dataset, i.e. the $r_{\mathbf x, \mathbf a}$ value. The only metric functions that are applicable here are $F_3$ 
and $F_4$, since these are the only ones that are heterogeneous, i.e. they can compute a similarity between a dataset and
a workflow. Note also that the Euclidean baseline strategy is no longer applicable because this can only be used between objects
of the same type. When it comes to the mean average error $F_3$ has a very poor performance compared to the default strategy. 
$F_4$ has a considerably better performance than $F_3$, thus providing further support to the incorporation of the additional 
constraints in the objective function, nevertheless this performance still is significantly worse than the default that of 
the default strategy.

\note[Alexandros]{I repeat something that I think I said previously, and which 
you also discuss bellow. We shoud try to get a better definition of 
the $\mathbf R$ matrix, or at least exploit its values in a more meaningful manner, cause as it is they are not
comparable between different pairs of datasets workflows.}

\note[Removed]{
For the third meta-mining task, we defined the parameters of the related objective functions as follows. 
$F_3$: $\mu_1 = 10$, $F_4$: $\alpha=1e^{-10}$, $\beta=1e^{-3}$, $\gamma=1e^{-3}$, $\mu_1 = 10$,  $\mu_2 = 0$.
We followed the same prior knowledge than in task 1 by giving more importance on the workflow patterns than on the dataset characteristics.
However these assumptions are potentially wrong since we have now to predict the goodness of match between a given dataset and a given workflow.
Regarding the two previous tasks, task 3 implies to consider datasets and workflows as equally informative 
such that we can derive a full order over dataset-workflow preferences. 
These preference scores should have an uniform distribution centered around a common mean value between datasets and workflows such that both sides of the preference matrix will have same variance.
But since we did not consider any normalization aspect in the setting of the preference matrix $\mathbf{R}$, 
we experimented this task with our own beliefs.

The results of task 3 are given in the right table of Table \ref{table:bio} where we did not get any significant results.
We have even a higher $\mae$ measure than the baseline for the two objective functions $F_3$ and $F_4$.  
This shows that task 3 is the most difficult meta-mining task compared to the two others,  
and we did not make any positive differences with the default score {\em def}. 
This happens because as said before we did not normalize correctly the preference scores. 
Thus the predicted preference for a given pair of dataset-workflow was not defined in a correct scale and gaves rise to predictions which fell outside the "normal" preference scale. 

Overall we get systematically better results with our heterogenous metric approach than the baseline {\em def} and the Euclidean metric {\em EC}. 
These results suggest that when the meta-data of the given test sample are not reliable enough, 
advising a heterogenous metric will greatly help over homogenous ones. 
This was the case in task 1. 
Otherwise, we get comparable results with homogenous metrics learned or defined in the Euclidean space. 
This was the case in task 2. 
In addition our heterogenous approach is made for optimizing the correlation between datasets/workflows. 
In tasks 1 and 2, we get systematically the best correlation measure with this approach with respect to the baseline and the homogenous methods. 
From a meta-mining perspective, 
improving the correlation $\rho$ in the dataset/workflow preferences is preferable than improving the mean average error, $\mae$, of these preferences 
because we would like to suggest the best performing workflow/dataset for the given meta-mining task.
Finally, we do not achieve significant results in task 3. 


}

Overall we tested a number of new metric-learning-based algorithms to solve different variants of the meta-mining 
problem following a hybrid recommendation approach. We have two metric-based-learning flavors, the homogeneous 
and the heterogeneous. In the homogeneous flavor we learn a metric in the original space in which some objects
are described, here datasets or workflows, which tries to approximate a similarity defined over a different space
that of the relative preference vectors. In the heterogeneous approach we learn a metric over the two different 
spaces that tries to reflect directly the goodness of match between the different objects. As it turns out the best
approach comes from the appropriate regularization of the heterogeneous metric by exploiting the 
additional constrains imposed on each of the original object spaces. In other words we seek for an heterogeneous 
metric defined over a common projection space of datasets and workflows, where the projection matrices of the datasets
and workflows are constrained to reflect vector preference similarities. In the immediate future we want to evaluate
the performance of the approach we presented here in recommendations problems other than the meta-mining with similar
problem requirements.

\section{Related Work}
\label{sec:rworks}

\note[Removed]{
In this section, we will discuss related works. 
These can be found mainly in the field of hybrid recommender systems~\cite{Agarwal2010,Agarwal2009,Stern2009,Burke2002}, 
where collaborative~\cite{Adomavicius2005,Su2009} and content-based filtering approaches~\cite{Pazzani_Billsus_2007} are combined together in order to improve item's recommendation to users. 
Hybrid approaches follow hence the same assumption we have for our meta-mining approach that similar users will generally prefer similar items.
Most related approaches are those that aim at solving the {\em cold-start} problem for users and/or items 
by adding users and items meta-data in a probabilistic factor-based model of the preference matrix~\cite{Agarwal2010,Agarwal2009,Stern2009}.  
Examples of recommender meta-data can be found for instance on the MovieLens dataset~\footnote{http://www.grouplens.org/}, 
and they concern demographic or activity information on users such as age, gender and occupation, and taxonomic information on movies such as genre and release date. 

Recommender meta-data differ however from meta-mining meta-data by their small number of descriptors.  
In recommendation systems, they are usually added in the factorization models under the form of Gaussian distributions~\cite{Stern2009} or Latent Dirichlet Allocation~\cite{Agarwal2010} to regularize these models. 
Thus, these approaches model side information by assuming that they are shaped in some normal distribution, and they learn user and item latent factors in the standard Euclidean space directly from the preference matrix. 
In our metric approach, we differ from these works by constraining the preference matrix with the additional information of the dataset and workflow similarities, learning then an appropriate heterogenous similarity measure 
which will be helpful when the available meta-data are not strongly reliable. 
We think thus that our metric approach might work not only on meta-mining problems but also on recommendation problems such as the MovieLens dataset  
since demographic, activity or taxonomic information are yet poor information. 
}

The meta-mining problem formulation we gave here is closely related with the work of hybrid
recommender systems~\cite{Agarwal2010,Agarwal2009,Stern2009,Burke2002}. There the goal is to
accurately recommend items to users using information on historical user preferences and descriptors of 
items and users.  Examples of recommender user and item descriptors can be found for instance on the 
MovieLens dataset where we have demographic or activity information on users such as age, gender and 
occupation, and taxonomic information on movies such as genre and release date. 

State of the art recommender methods~\cite{Agarwal2010,Agarwal2009,Stern2009}, 
rely on matrix factorization methods  to directly approximate the preference matrix 
as we do in the optimization problem~(\ref{eq:metric-ds-wf.2}). 
In \cite{Stern2009}, the authors proposed a Bayesian approach where a probabilistic bi-linear rating model 
is inferred by a combination of expectation propagation and variational message passing. 
Users and items features are modelled with Gaussian priors into two matrices 
$\mathbf{U}$ and $\mathbf{V}$ of latent {\em traits}, the inner product 
of which defines user-item similarities. Variational approximation on 
users and items is then used to regularize the latent factors.   
Their experiments on the MovieLens dataset showed that including user and item descriptors 
improves performance. 
\note[Removed]{
, especially for high values of $K$.  }
\note[Alexandros]{Which brings me to a question that I have not thought before. How do you control/set
the dimensionality $t$ of the common projection space? We say nothing about that in the paper.}
\cite{Agarwal2010,Agarwal2009}, propose also a generative probabilistic model where the 
model fitting is done by a Monte Carlo EM algorithm with no variational approximations. 
They regularize their model using a regression-based approach on user and item factors, 
where the latter are determined using topic modelling,~\cite{Agarwal2010}. 
They also experimented on the MovieLens dataset and showed that 
a model based on the  meta-data had a weak predictive performance
while their regression-based approach to the latent factors regularization gave the best 
performance improvements. 
Our metric-learning-based approach to the problem of hybrid recommendation uses a very different
regularization approach in learning the factorization matrices, we
focus on constraining them in a manner that they will reflect in the original 
feature spaces the similarities of the respective preference vectors, approach
which in our meta-mining experiments had the best performance.
One additional advantage of the use of the two linear projection matrices learned
in the dataset and workflow spaces is that we can now naturally handle the out-of-sample 
problem, i.e. the cold-start problem in recommender systems, which is not the case with
the typical matrix factorization models.

\section{Conclusion and Future Work}
\label{sec:conc}

In this paper we take a new view on the relatively new concept of meta-mining, view that is also relevant for the 
more traditional work of meta-learning. We model the problem of the selection of the appropriate workflow or algorithm
for a dataset as a hybrid recommendation problem, in which suggestions will be provided based on the descriptors
of the dataset and the workflow or algorithm. To that end we propose a new metric-learning-based approach to the
hybrid recommendation problem, which learns homogeneous metrics in the original dataset and workflow spaces,
constrained in a manner that will reflect workflow preference and dataset preference vector 
similarities, and combines them with an heterogeneous metric in the dataset-workflow space 
that reflects the appropriateness of a given workflow for a given dataset. The two 
homogeneous metric-learning problems act as additional, relevant, regularizers for the heterogeneous
metric learning problem. In addition thanks to the linear projections that lie at the core of our 
method, it is able to handle in a natural manner the cold-start problem.
The combined use of the three metrics achieves the best results. To the best of our knowledge this 
the first approach of its kind, not only for the 
meta-mining problem, but as well as for the more general problem of the hybrid recommendation. Our immediate
goal is to experiment with our approach to standard hybrid recommendation problems, such as the MovieLens dataset, 
and compare its performance with typical recommendation approaches used in such problems.

\note[Removed]{
In this paper, we have seen a metric-learning approach for hybrid recommendations in meta-mining with strong focus on the available meta-data. 
To our knowledge, our approach is the first of its kind and can be applied in a variety of cold-start recommendation problems. 
Our approach also contribute to the meta-mining framework by reformulating it as a matching problem between datasets and workflows which is more appropriate than the usual classification or regression-based approaches. 
We demonstrated 
our approach on a real-world meta-mining problem  
for the three meta-mining tasks of preference learning. 
We achieved systematically and significantly better results 
for the first two tasks. 
In future works, we intend to 
validate our approach against classification-based meta-mining approaches such as the works of \cite{Hilario2010}. 
As well, we intend to test our approach on recommendation problems such as the MovieLens dataset. }


\bibliographystyle{IEEEtran}
\bibliography{ref}

\end{document}